\documentclass{article} 
\usepackage{iclr2027_conference,times}
\iclrfinalcopy  

\usepackage{amsmath,amsfonts,bm}

\def\eqref#1{equation~\ref{#1}}

\def\1{\bm{1}}

\DeclareMathAlphabet{\mathsfit}{\encodingdefault}{\sfdefault}{m}{sl}
\SetMathAlphabet{\mathsfit}{bold}{\encodingdefault}{\sfdefault}{bx}{n}

\usepackage{hyperref}
\usepackage{url}
\newcommand{\method}{\texttt{TICDA}}
\usepackage{amsfonts} 
\usepackage{amssymb}
\usepackage{bm}
\usepackage{tcolorbox}
\usepackage{xcolor}
\usepackage{enumitem}
\usepackage{bm}
\usepackage{booktabs}
\usepackage{multirow}
\usepackage{subcaption}
\usepackage{lipsum}
\usepackage{wrapfig}
\usepackage{float}
\usepackage{etoc}
\usepackage{subcaption} 
\usepackage{booktabs}
\usepackage{pifont}
\usepackage{soul}
\usepackage{xcolor}
\usepackage{soul}
\usepackage{wrapfig}
\usepackage{graphicx}
\usepackage{float}
\usepackage{placeins}
\usepackage{subcaption}  

\sethlcolor{yellow}

\title{\method: Tabular In-Context Data Attribution}

\author{
Yacine Benihaddadene$^{1,*}$ \And
Milan Bhan$^{1,*}$ \And
Eliot Dugelay$^{2}$ \AND
Mohammed Jawhar$^{1}$ \And
Benjamin Wong$^{1}$ \And
Nicolas Chesneau \And
Duong Nguyen \\[1.5ex]
$^{1}$Ekimetrics \quad
$^{2}$ETH Zurich \\[0.5ex]
$^{*}$Equal contribution
}

\providecommand{\TICDA}{\texttt{TICDA}}
\providecommand{\DETAIL}{\texttt{DETAIL}}

\begin{document}

\maketitle

\begin{abstract}

Tabular foundation models (TFMs) achieve strong predictive performance by conditioning on labeled demonstrations provided in context, without any parameter update. Yet how individual demonstrations shape a given prediction remains poorly understood. This gap matters in practice: the context is often assembled from whatever labeled data is available, potentially leading to the inclusion of mislabeled, redundant, or low-quality examples that degrade performance. Standard data attribution methods do not transfer to the TFM setting: resampling-based approaches such as \texttt{DemoShapley} require a combinatorial number of forward passes, and gradient-based estimators such as influence functions require computing training point's effect on the model parameters, which in-context learning never updates. We introduce \method, a method that measures the influence of every demonstration in the context directly from linear surrogates trained on TFM latent embeddings, in a single forward pass and at negligible cost. We show that \method\ offers the best compromise against competitors across four tasks: detecting labeling errors, curating context to preserve predictive accuracy while lowering inference cost, producing attribution scores that transfer across TFMs, and supporting an acquisition strategy for efficient active learning.

\end{abstract}

\section{Introduction}

Tabular foundation models (TFMs) have demonstrated that transformer-based architectures using in-context learning (ICL)~\citep{dong2024survey} can achieve state-of-the-art
performance on small- to medium-sized predictive tabular tasks~\citep{erickson2026tabarena,purucker2026beyond}. A TFM produces a prediction by conditioning on a set of labeled demonstrations in its context, without any parameter update. A succession of models has extended this paradigm to larger and more heterogeneous datasets~\citep{hollmann2023tabpfn, qu2025tabicl, hollmann2025tabpfn, qu2026tabiclv2, ma2024tabdpt}, raising the interest in TFMs~\citep{vanbreugel2024position}. Despite this success, comparatively little is understood about how TFMs use their context, and in particular about the influence that each individual demonstration exerts on a given prediction~\citep{rundel2024interpretable}. This influence is not merely of interpretive interest: since the context is assembled from whatever labeled data happens to be available, mislabeled, redundant, or otherwise low-quality demonstrations can be introduced silently and either degrade predictive performance without any obvious signal or induce useless computational overhead. Those effects have been observed in TFMs: TabPFN loses accuracy when its demonstrations
come from a distribution that drifts over time \citep{helli2024drift}, unsupervised
extensions of TabPFN to anomaly detection become unstable when their context is noisy
or contaminated \citep{marszalek2026tactic}, and \citet{rundel2024interpretable} show that a TabPFN context chosen by data valuation predicts better than a random context of the same size, which matters
because the cost of TabPFN grows quadratically with the number of demonstrations. The combination of the widespread deployment of TFMs and the opacity of their in-context inference makes attributing predictions to individual demonstrations increasingly critical.

\emph{Training data attribution} (TDA) methods aim to quantify the influence that individual training points exert on a model's behavior and performance~\citep{hammoudeh2024training}. Transposing this idea to TFMs is not straightforward: the relevant unit is a demonstration conditioning a frozen pretrained model through its context, rather than a training point that shaped a set of learned parameters. Gradient-based estimators such as influence functions~\citep{koh2017influence} are defined through the model parameters and thus have no natural counterpart for a demonstration whose effect is exerted at inference time. Gradient-free resampling-based approaches such as \texttt{Data Shapley}~\citep{ghorbani2019datashapley} require evaluating the model over a combinatorial number of data subsets, which is prohibitive when each evaluation amounts to a full forward pass over the context. To overcome these limitations, we introduce \method, a gradient-based approach for estimating the influence of every demonstration in the context at negligible cost.

Our main contributions
are as follows:

\begin{enumerate}[noitemsep,topsep=0pt,leftmargin=*]
    \item We propose \method, which measures the influence of every in-context demonstration directly from linear surrogates trained on TFM latent embeddings, in a single forward pass.
    \item We show that \method\ detects labeling errors among demonstrations competitively with existing methods, while incurring lower cost.
    \item We show that \method\ enables context curation that either improves or maintains predictive performance while reducing inference cost.
    \item We establish that the attribution scores produced by \method\ transfer across heavier TFMs.
    \item We derive an acquisition strategy from \method\ influence scores for efficient active learning.
\end{enumerate}

This paper is organized as follows.
Section~\ref{ticda:method} introduces \method, and Section~\ref{sec:experimental_setting} shows how
\method\ enables accurate labeling error detection, context curation,
TFM transfer and active learning. Section~\ref{sec:related-work} gives a review of
how existing approaches compute data attribution, both for deep models
in general and in the ICL setting.

\section{TICDA: A Method to Compute fixed-representation influences}
\label{ticda:method}

In this section, we introduce \method, a method for measuring the
influence that each demonstration of a TFM context exerts on a given
prediction. We measure this influence as a leave-one-out (LOO) effect: the change of the loss on a labeled query $z_q$ when demonstration $i$ is removed and the prediction is made again without it, where $z_q=(x_q,y_q)$ is an input $x_q$ with its label $y_q$.
Let $\mathcal{D}=\{(x_j,y_j)\}_{j=1}^{n}$ be a classification context with
$K$ classes and observed one-hot labels $y_j\in\mathbb{R}^{K}$, let $f$
be a frozen TFM, and write $z_j=(x_j,y_j)$ for the $j$-th demonstration.
The LOO influence of demonstration $z_i$ on the query $z_q$ is the
change in the loss of the TFM when the demonstration is removed from the
context:
\begin{equation}
 \Delta^{\mathrm{TFM}}_{iq}
 =\mathcal{L}\big(f(x_q;\mathcal{D}_{\setminus i}),\,y_q\big)
 -\mathcal{L}\big(f(x_q;\mathcal{D}),\,y_q\big),
 \label{ticda:loo}
\end{equation}
where $f(x_q;\mathcal{D})$ is the prediction of the TFM for $x_q$ given the context $\mathcal{D}$, $\mathcal{D}_{\setminus i}$ is the context without demonstration $i$, and $\mathcal{L}$ 
is the predictive loss. A positive value of $\Delta^{\mathrm{TFM}}_{iq}$ means that
the demonstration helps the query, since its removal raises the loss. Evaluating
Equation~\ref{ticda:loo} for every demonstration requires $n$
forward passes over contexts of $n-1$ demonstrations, which is
prohibitive for the contexts of thousands of demonstrations that TFMs
are designed for.

To avoid doing one forward pass per demonstration, we follow the influence function approach
\citep{koh2017influence}.
Consider a model with parameters $\theta$ (pre)trained on the $n$ demonstrations of
$\mathcal{D}$: its fitted parameters $\widehat{\theta}$ minimize the
empirical risk $\frac{1}{n}\sum_{j=1}^{n}\mathcal{L}(z_j,\theta)$, with
$\mathcal{L}(z_j,\theta)$ the loss of its prediction on $x_j$ against
$y_j$. Upweighting a training point $z_i$ by an infinitesimal amount
changes the loss at $z_q$ by
\begin{equation}
 \mathcal{I}_{\mathrm{up,loss}}(z_i,z_q)
 = -\nabla_{\theta}\mathcal{L}(z_q,\widehat{\theta})^{\top}
   H_{\widehat{\theta}}^{-1}\,
   \nabla_{\theta}\mathcal{L}(z_i,\widehat{\theta}),
 \label{eq:influence}
\end{equation}
where $\nabla_{\theta}$ denotes the gradient with respect to the
parameters and
$H_{\widehat{\theta}}=\frac{1}{n}\sum_{j=1}^{n}\nabla^{2}_{\theta}\mathcal{L}(z_j,\widehat{\theta})$
is the Hessian. A positive value of $\mathcal{I}_{\mathrm{up,loss}}(z_i,z_q)$
means that giving the training point $z_i$ more weight raises the loss at
$z_q$, and vice versa. $\mathcal{I}_{\mathrm{up,loss}}(z_i,z_q)$ is the
first-order estimate (up to a factor) of the change of $\Delta^{\mathrm{TFM}}_{iq}$
\citep{koh2017influence}.

The influence function approach relies on differentiating the loss with respect to the model's parameters $\theta$, through which a training point exerts its effect.
However, a TFM learns in context, and its parameters remain fixed during
inference. A demonstration influences a prediction only through the
context in which the model reads it and there are no parameter updates. 
Hence, the gradients required in Equation~\ref{eq:influence} cannot be computed.

\begin{figure*}[t]{\centering}
\begin{center}
\includegraphics[width=\linewidth]{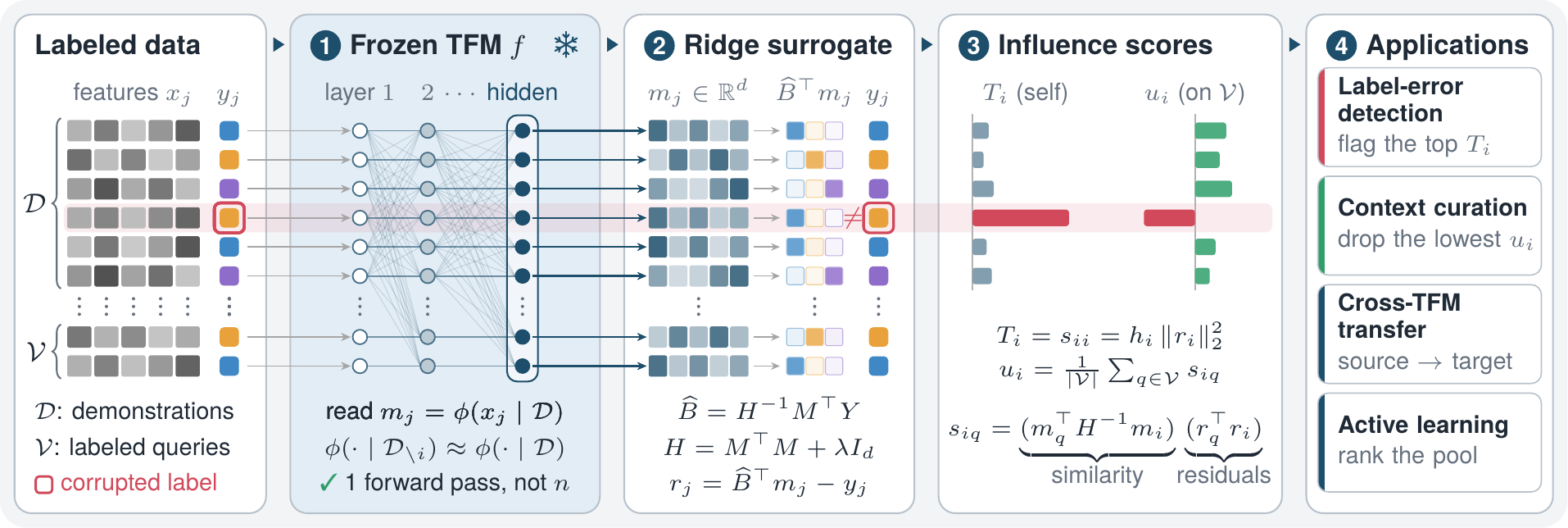}
\caption{\method\ overview. \method\ (1) extracts TFM embeddings, (2) trains a linear surrogate on the classification task of interest, (3) estimates influence score based on the surrogate and (4) can be applied for label-error detection, context curation, cross-TFM transfer and active learning.}  
\label{fig:overall}
\end{center}
\end{figure*}

In order to overcome this limitation, \method\ operates in two steps: (1)~\method\ fits a ridge surrogate on the representations that the TFM constructs for the demonstrations (Section~\ref{ticda:target}), and (2)~\method\ holds these representations fixed (Section~\ref{ticda:fixedrep}). Step (1)~provides the parameters that influence functions require, while step (2)~allows a single forward pass to serve all demonstrations. Under these two choices, Equation~\ref{eq:influence} admits a closed-form solution yielding two scores, self-influence and query-influence (Section~\ref{ticda:influence}). Figure~\ref{fig:overall} gives an overview of the \method\ methodology and we detail its components in the following sections.

\subsection{Attribution surrogate}
\label{ticda:target}

To obtain parameters to which every demonstration contributes, we propose to apply Equation~\ref{eq:influence} to a  ridge linear probe fitted on the representations of the TFM rather than to its weights. Let $\phi(\cdot\mid\mathcal{D})\in\mathbb{R}^{d}$ denote the row representation read at a given hidden layer of $f$ when the context is $\mathcal{D}$. We write $m_j=\phi(x_j\mid\mathcal{D})$ for
demonstration $j$, $m_q=\phi(x_q\mid\mathcal{D})$ for a query, and $M\in\mathbb{R}^{n\times d}$ and $Y\in\mathbb{R}^{n\times K}$ for the demonstration and label matrices. This notation makes the dependence on $\mathcal{D}$ explicit: in a TFM, the representation of a row is
computed from the whole context, including its labels for the backbones
that we study, and not from the row alone. On these representations, we
fit a ridge surrogate to the observed labels:
\begin{equation}
 \widehat{B}=\arg\min_{B}J(B),\qquad
 J(B)=\frac12\lVert MB-Y\rVert_F^{2}
      +\frac{\lambda}{2}\lVert B\rVert_F^{2},
 \label{ticda:objective}
\end{equation}
where $\|\cdot\|_F$ is the Frobenius norm and $\lambda>0$ is the regularization term.
This objective admits the closed-form solution
\begin{equation}
 \widehat{B}=H^{-1}M^{\top}Y,\qquad
 H=M^{\top}M+\lambda I_d.
 \label{ticda:fit}
\end{equation}
For a row with representation $m_j$, the surrogate predicts the class-score vector $\widehat{B}^{\top}m_j$, its residual $r_j=\widehat{B}^{\top}m_j-y_j$ measures the discrepancy from the observed one-hot label, and $\ell_j(B)=\ell(B^{\top}m_j,y_j)$ denotes its loss, with $\ell(s,y)=\tfrac12\lVert s-y\rVert_2^{2}$. The surrogate provides parameters, $\widehat{B}$, to which every demonstration contributes and through which Equation~\ref{eq:influence} becomes
applicable. 

\subsection{Fixed representations simplification}
\label{ticda:fixedrep}

The surrogate alone does not remove the cost of
Equation~\ref{ticda:loo}. Removing demonstration $i$ changes the
surrogate loss on the query by
\begin{equation}
 \Delta^{\mathrm{R}}_{iq}
 =\ell\big(\widehat{B}(\mathcal{D}_{\setminus i})^{\top}
           \phi(x_q\mid\mathcal{D}_{\setminus i}),\,y_q\big)
 -\ell\big(\widehat{B}(\mathcal{D})^{\top}
           \phi(x_q\mid\mathcal{D}),\,y_q\big),
 \label{ticda:recomputed}
\end{equation}
where $\widehat{B}(\mathcal{D})$ denotes the coefficients fitted on the context $\mathcal{D}$, and this change has two distinct origins: (1)~demonstration $i$ leaves the regression, so that the coefficients are refitted, which amounts to a rank-one update of a $d\times d$ system, and (2)~the context itself changes, so that the TFM re-encodes every remaining demonstration and the query,
$M(\mathcal{D}_{\setminus i})$ being different from $M(\mathcal{D})$ with one row removed. The second effect requires a forward pass of the TFM for each candidate deletion. Equation~\ref{ticda:recomputed} is thus as expensive as Equation~\ref{ticda:loo}, and an approach willing to pay this cost would have no use for a surrogate, since it could measure the loss of the TFM directly.

Motivated by the fact that one demonstration represents a small fraction of a context of thousands of demonstrations, we make the following simplification: for every demonstration $i$,
\begin{equation}
 \phi(\cdot\mid\mathcal{D}_{\setminus i})
 \;\approx\;\phi(\cdot\mid\mathcal{D}).
 \label{ticda:fixed}
\end{equation}

We provide evidence in Appendix~\ref{ticda:app-fixed-measure} that Equation~\ref{ticda:fixed} is reasonable. Specifically, differences are approximately $0.5\%$ of the representation norm with 4,096 demonstrations and decrease as the number of demonstrations increases. All representations are extracted once, from the full context, and the removal of a demonstration only affects the fit of the surrogate. This choice has two consequences. First, a single forward pass serves all $n$ demonstrations instead of one pass each. Second, the surrogate becomes an ordinary ridge regression on a fixed design, i.e., a strictly convex and twice-differentiable objective. That is the setting in which the influence function of Equation~\ref{eq:influence} holds. 
\subsection{Demonstration upweighting influence}
\label{ticda:influence}

Under Equation~\ref{ticda:fixed}, we follow \citet{koh2017influence} and measure a demonstration's influence by slightly modifying its weight in the ridge objective and evaluating the resulting change in query loss. Changing the weight of demonstration $i$ from $1$ to $1+\epsilon$
gives
\begin{equation}
 \widehat{B}_\epsilon
 =\arg\min_{B}\big[J(B)+\epsilon\,\ell_i(B)\big],
 \label{ticda:perturbation}
\end{equation}
where the representations and the global regularization penalty remain fixed. Differentiating the normal equations at $\epsilon=0$ gives $d\widehat{B}_\epsilon/d\epsilon=-H^{-1}m_i r_i^{\top}$. No Hessian over the parameters of the TFM is involved. We define the influence of demonstration $i$ on query $q$ as
\begin{equation}
 s_{iq}:=-\left.
   \frac{d\,\ell_q(\widehat{B}_\epsilon)}{d\epsilon}
   \right|_{\epsilon=0}
 =\big(m_q^{\top}H^{-1}m_i\big)\big(r_q^{\top}r_i\big).
 \label{ticda:support}
\end{equation}
The sign indicates whether the demonstration locally helps or hurts the query under the surrogate: $s_{iq}>0$ means that increasing its weight reduces query loss, while $s_{iq}<0$ means that it increases query loss. The pairwise score $s_{iq}$ describes the influence of demonstration $i$ on the surrogate loss at a query $q$. Evaluating it on the demonstration itself or aggregating it over several queries yields two complementary summaries, which can be used for different applications.




\paragraph{Self-influence.} Using demonstration $i$'s own representation and label as the query in Equation~\ref{ticda:support} gives its self-influence: $ T_{i}=s_{ii}=h_i\,\lVert r_i\rVert_2^{2}$.
Defined this way, $T_{i}$ is the first-order change of the demonstration's own loss when it is removed. This score is the squared residual multiplied by the leverage $h_i$ (Appendix~\ref{ticda:app-first-order}). 
Demonstrations with a large $T_{i}$ are those whose observed label is not supported by the rest of the context. Hence, $T_{i}$ can be used to detect labeling errors, as we show in Section~\ref{sec:experimental_setting}.

\paragraph{Query-influence.}  Given a set $\mathcal{V}$ of queries, the mean influence $u_i=\frac{1}{|\mathcal{V}|}\sum_{q\in\mathcal{V}}s_{iq}$
shows how demonstration $i$ supports the predictions on $\mathcal{V}$. Demonstrations with a negative $u_i$ are those whose removal lowers the mean loss on $\mathcal{V}$. Hence, $u_i$ can be used to
curate the context, by removing the demonstrations with the lowest $u_i$ computed on a labeled validation set (Section~\ref{sec:experimental_setting}).

\section{Experimental Settings}
\label{sec:experimental_setting}



This section presents the experimental study
conducted across 38 datasets from TabArena~\citep{erickson2026tabarena} and several TFMs. We first run \method\ on two TFMs to assess how well it detects labeling errors, including a sensitivity analysis over the hidden space size and the influence computation method (self or query). We then show how \method\ enables to reduce context size, and therefore the inference cost of TFMs, without sacrificing performance, even improving it in some cases. Finally, we show how \method\ can be used to build an acquisition strategy for active learning under low data regime.

\subsection{Experimental Setup}
\label{sec:experimental_protocol}

\paragraph{Datasets.}
\method\ and competitors are tested on 38 classification datasets from TabArena. We follow the preprocessing and evaluation protocol of TabArena and limit each context to at most \(4{,}096\) demonstrations. All methods use the same context and test splits on each dataset.


\paragraph{Models.}
We consider two frozen TFMs as backbone for \method: TabICLv2~\citep{qu2026tabiclv2} and TabDPT 1.2~\citep{ma2024tabdpt}. \method\ reads the row representations of TabICLv2 before its in-context learning layers and those of TabDPT after the 16th of its 32 transformer layers.
No backbone parameters are updated during attribution. 
We first evaluate each attribution method independently on every backbone. Then, to test whether the scores transfer to models that TICDA never reads, we rank the demonstrations with one of these two backbones and apply this ranking unchanged to curate the context of TabPFN-3~\citep{grinsztajn2026tabpfn3} and TabPFN-3.5~\citep{jager2026tabpfn35}.

\paragraph{Evaluation Protocols and Metrics.}

We evaluate \method\ through \texttt{LOO} estimation, labeling error detection, computational efficiency, context curation, transfer to other TFMs, and acquisition strategy for active learning. For \emph{\texttt{LOO} estimation}, we compute the Spearman correlation between \method\ and exhaustive leave-one-out (\texttt{LOO}), as defined in Equation~\ref{ticda:loo}, and refer to this agreement as faithfulness. 
For \emph{labeling error detection}, we follow~\cite{ghorbani2019datashapley}: we corrupt the labels of \(20\%\) of the context demonstrations, treat them as positives, rank demonstrations by self-influence ($T_i$), and report the AUC of the recovered ranking. We report attribution runtime to assess \emph{computational efficiency}. For \emph{context curation}, under a fixed budget we rank demonstrations by their mean influence ($u_i$) on a validation set, remove the lowest-scored, and re-evaluate the backbone on the unchanged test set, comparing balanced accuracy against the uncurated context and against random removal. For \emph{active learning}, we test whether \method\ can guide data acquisition in a pool-based setting by acquiring the candidates whose addition most reduces the surrogate loss on a labeled validation set, and measure label efficiency through the area under the learning curve.

\paragraph{\method\ and Competitors.}
We use the base configuration of \method\ introduced in Section~\ref{ticda:method}. Given the frozen row representations, \method\ fits a ridge-regularized linear probe and derives the self-influence scores ($T_{i}$) used for label-error detection and the mean query-influence scores ($u_i$) used for context curation without further calls to the backbone. We fix the ridge regularization coefficient to \(\lambda=10\) for all datasets and backbones, without dataset-specific tuning. We also implement \texttt{TICDA-IT} for context curation experiments, consisting in performing iterative curation, where $5\%$ of the demonstrations are removed at each step, followed by refitting and re-ranking with the \method\ data attribution computation protocol, until the target removal threshold is reached. We compare \method\ to four competitors. We also run Integrated gradients (\texttt{IG})~\citep{sundararajan2017axiomatic} to estimate each demonstration's contribution by integrating prediction gradients along a path from a reference input, constituted by replacing the input demonstration of interest with padding. We implement other competitors that we present in more details in Section~\ref{sec:related-work}:  \texttt{DETAIL}~\citep{zhou2024detail} (following basic setting from the paper) and an affordable version of \texttt{DemoShapley}~\citep{xie2025demoshapley} for TFMs to approximate Shapley values from eight Monte Carlo context permutations. Beyond being used as a reference to compute faithfulness, we also compare our results to Leave-one-out (\texttt{LOO}) for labeling error detection and context curation.



\begin{table}[t]
\caption{Data attribution for in-context demonstrations on tabular foundation models.
Results are aggregated over 38 datasets (mean $\pm$ std). Best results are
\textbf{bold}, second best \underline{underlined}; stars compare the winner with the
runner-up (two-sided paired $t$-test, $^{*}p<.10$, $^{**}p<.05$,
$^{***}p<.01$).}
\label{tab:main-results}
\centering
\footnotesize
\setlength{\tabcolsep}{4pt}
\resizebox{\linewidth}{!}{%
\begin{tabular}{llccccc}
\toprule
& & \texttt{LOO} & \texttt{DemoShapley} & \texttt{IG} & \texttt{DETAIL} & \method\ (ours) \\
\midrule
\multirow{2}{*}{\textbf{TabDPT}}
 & Faith. $\uparrow$ & --- & $\mathbf{0.91}^{***}$ {\scriptsize $\pm 0.07$} & $0.20$ {\scriptsize $\pm 0.35$} & $-0.06$ {\scriptsize $\pm 0.18$} & $\underline{0.79}$ {\scriptsize $\pm 0.09$} \\
 & AUC $\uparrow$   & $\mathbf{0.90}^{***}$ {\scriptsize $\pm 0.08$} & $\underline{0.88}$ {\scriptsize $\pm 0.10$} & $0.60$ {\scriptsize $\pm 0.20$} & $0.47$ {\scriptsize $\pm 0.13$} & $\underline{0.88}$ {\scriptsize $\pm 0.08$} \\
\midrule
\multirow{2}{*}{\textbf{TabICL}}
 & Faith. $\uparrow$ & --- & $\underline{0.65}$ {\scriptsize $\pm 0.15$} & $0.30$ {\scriptsize $\pm 0.22$} & $-0.11$ {\scriptsize $\pm 0.22$} & $\mathbf{0.67}$ {\scriptsize $\pm 0.21$} \\
 & AUC $\uparrow$   & $\underline{0.80}$ {\scriptsize $\pm 0.16$} & $0.76$ {\scriptsize $\pm 0.11$} & $0.64$ {\scriptsize $\pm 0.18$} & $0.37$ {\scriptsize $\pm 0.17$} & $\mathbf{0.89}^{***}$ {\scriptsize $\pm 0.08$} \\
\bottomrule
\end{tabular}%
}
\end{table}

\subsection{Experimental Results}

\paragraph{\texttt{LOO} Faithfulness and Labeling Error Detection.}

Table~\ref{tab:main-results} reports results aggregated over the 38 datasets from TabArena. Across both backbones, \method\ ranks first or second on every metric. On TabDPT, \method\ is the second-best method for both faithfulness ($0.79$) and AUC ($0.88$), trailing \texttt{DemoShapley} on faithfulness and tying it on labeling error detection. On TabICL, \method\ reaches the best faithfulness ($0.67$) and the best AUC ($0.89$), the latter significantly above \texttt{LOO}. No competitor is this consistent: \texttt{DemoShapley} is strong on TabDPT but degrades on TabICL, while \texttt{IG} and \texttt{DETAIL} lead to poor results both in terms of faithfulness and labeling error detection. 

\begin{wrapfigure}{r}{0.50\textwidth}
    \centering
    \includegraphics[width=0.45\textwidth]{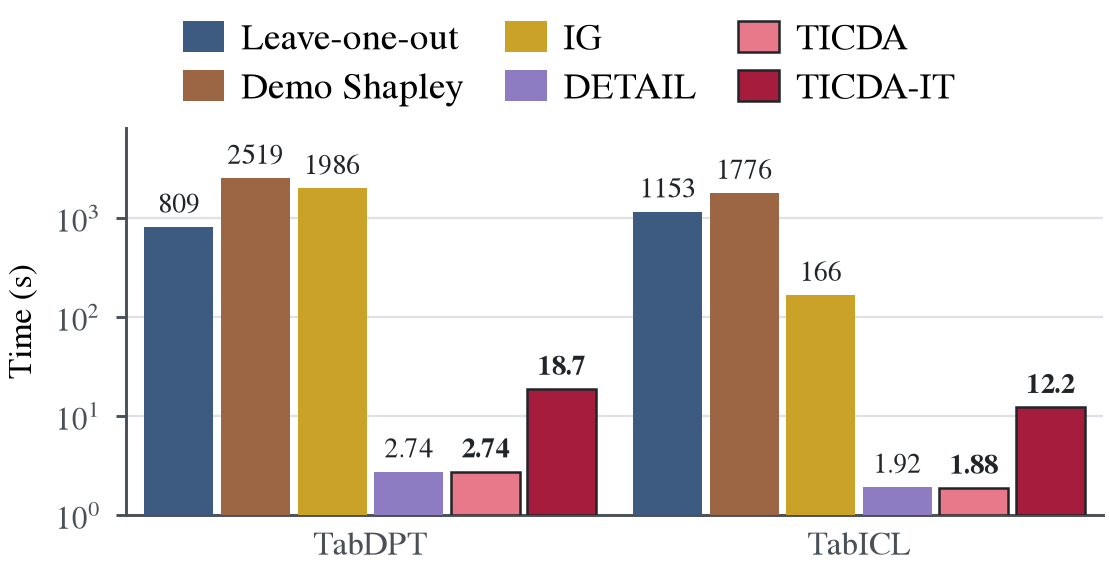}
    \caption{Average computation time per method.}
    \label{fig:compute_time}
\end{wrapfigure}
 
Figure~\ref{fig:compute_time} additionally reports computation time per method (log scale). \method\ and \texttt{DETAIL} are the only methods that run in under three seconds ($1.88$--$2.74$\,s), two to three orders of magnitude faster than \texttt{LOO}, \texttt{DemoShapley}, and \texttt{IG}. \texttt{DETAIL} is cheap but, as Table~\ref{tab:main-results} shows, performs poorly on faithfulness and labeling error detection. \method\ is thus the only method that pairs top-tier faithfulness and labeling-error detection with negligible cost, making it the best overall compromise.


We ablate two design choices of \method: the influence score used to rank demonstrations and the dimensionality of the representations on which attribution is performed. We compare self-influence \(T_{i}\) with query-influence, as defined in Section~\ref{ticda:influence}. Following \cite{zhou2024detail}, we also compare the full \(512\)-dimensional TabICL representations with a random projection to \(20\%\) of their original dimension, corresponding to \(102\) dimensions. Results for the reduced representations are averaged over three projection seeds. We report the same quality metrics as above: faithfulness with \texttt{LOO}, label-error detection performance (AUC), and computation runtime.

\begin{wraptable}{r}{0.50\textwidth}
\caption{Ablation study on \textbf{TabICL} (38 datasets, $20\%$ label corruption, mean $\pm$ std).}
\label{tab:ablation}
\centering
\small
\resizebox{0.48\textwidth}{!}{%
\begin{tabular}{ccccc}
\toprule
\multicolumn{2}{c}{\textbf{Configuration}} & \multicolumn{3}{c}{\textbf{Metrics}} \\
\cmidrule(lr){1-2} \cmidrule(lr){3-5}
Influence & Dimension & Faith. $\uparrow$ & AUC $\uparrow$ & Time (s) $\downarrow$ \\
\midrule
query & $20\%$ & $0.32_{\scriptscriptstyle \pm 0.16}$ & $0.85_{\scriptscriptstyle \pm 0.10}$ & $4.90_{\scriptscriptstyle \pm 1.30}$ \\
query & full   & $0.33_{\scriptscriptstyle \pm 0.16}$ & $0.86_{\scriptscriptstyle \pm 0.10}$ & $5.09_{\scriptscriptstyle \pm 1.31}$ \\
self  & $20\%$ & $0.67_{\scriptscriptstyle \pm 0.21}$ & $0.89_{\scriptscriptstyle \pm 0.08}$ & $3.51_{\scriptscriptstyle \pm 0.68}$ \\
\midrule
self  & full   & $0.67_{\scriptscriptstyle \pm 0.21}$ & $0.89_{\scriptscriptstyle \pm 0.08}$ & $3.71_{\scriptscriptstyle \pm 0.69}$ \\
\bottomrule
\end{tabular}%
}
\end{wraptable}

Table~\ref{tab:ablation} shows that the influence scoring methodology has a substantially larger effect than the representation dimensionality. With full representations, self-influence significantly increases faithfulness (from 0.33 to 0.67) and AUC (from 0.86 to 0.89) relative to query-influence. Reducing the representation dimension leaves AUC and faithfulness unchanged, while only slightly improving computation time. These results especially support the use of \method\ self-influence for label-error detection. We additionally show in Appendix~\ref{app:add_xp} the sensitivity of our approach to the regularization coefficient.

\begin{figure*}[t]{\centering}
\begin{center}
\includegraphics[width=0.90\linewidth]{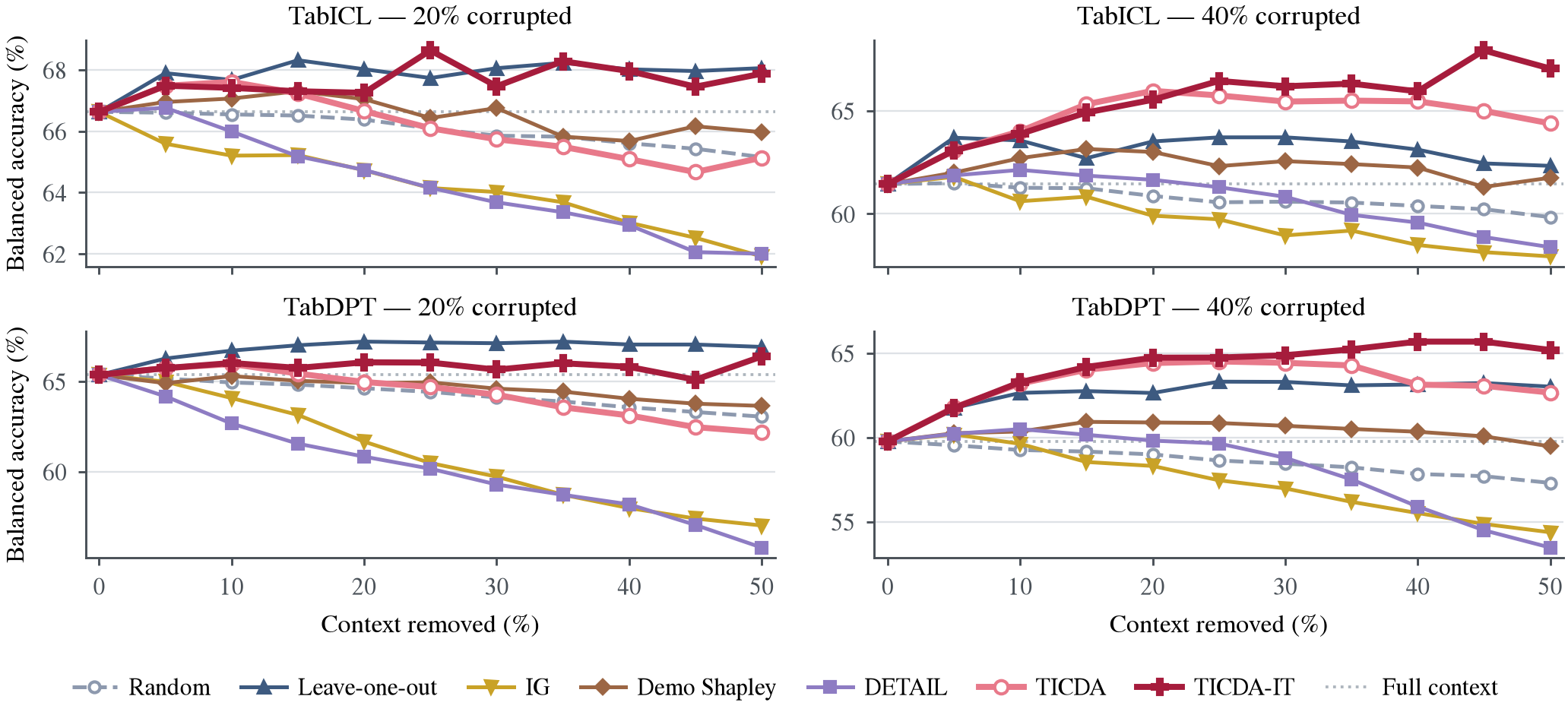}
\caption{Curation: aggregated balanced accuracy evolution on \textbf{TabDPT} and \textbf{TabICL} under different levels of demonstration corruption.
Each method ranks in-context demonstrations by its attribution scores, and
the lowest-scored demonstrations are removed.}  
\label{fig:curation}
\end{center}
\end{figure*}


\paragraph{Context Curation.}

Figure~\ref{fig:curation} reports aggregated curation results on the 38 selected TabArena datasets, at two initial corruption levels ($20\%$ and $40\%$) and across increasing fractions of removed context. \texttt{TICDA-IT} is the most robust curation strategy across both settings: it keeps balanced accuracy at or above the full-context level as demonstrations are removed, whereas \texttt{IG} and \texttt{DETAIL} degrade sharply, \texttt{DemoShapley} tracks the random baseline, and \texttt{LOO} curates well only at $20\%$ corruption. The gain is clearest under heavy corruption: at $40\%$, the balanced accuracy of both \method\ and \texttt{TICDA-IT} increases with removal and peaks around $40$--$45\%$, so that discarding up to half of the demonstrations outperforms the full context. Read together with the compute results of Figure~\ref{fig:compute_time}, this shows that \texttt{TICDA}-based curation preserves or improves accuracy while using up to $50\%$ fewer demonstrations, at a small fraction of the cost of \texttt{LOO}, the only baseline of comparable curation quality. Finally, \texttt{TICDA-IT} consistently outperforms \method, most visibly at $20\%$ corruption where \method\ alone drifts below the full-context level. Recomputing data attribution during curation is therefore worthwhile.

\begin{table}[t]
\caption{\method\ TDA Transfer for context curation between source and target models. Balanced accuracy (mean $\pm$ std, aggregated over all datasets) is reported at several levels of removal.}
\label{tab:transfer}
\centering
\small
\begin{tabular}{llcccc}
\toprule
& & \multicolumn{4}{c}{\textbf{Demonstrations removed}} \\
\cmidrule(lr){3-6}
Target & Source & $0\%$ & $10\%$ & $20\%$ & $50\%$ \\
\midrule
\multirow{3}{*}{TabPFN-3} & Random & $59.0$ {\scriptsize$\pm 14.9$} & $58.7$ {\scriptsize$\pm 14.6$} & $58.3$ {\scriptsize$\pm 14.4$} & $56.6$ {\scriptsize$\pm 14.1$} \\
& TabICLv2 & $59.0$ {\scriptsize$\pm 14.9$} & $\mathbf{63.5}$ {\scriptsize$\pm 15.8$} & $\mathbf{65.0}$ {\scriptsize$\pm 15.8$} & $\mathbf{64.4}$ {\scriptsize$\pm 15.4$} \\
& TabDPT & $59.0$ {\scriptsize$\pm 14.9$} & $62.6$ {\scriptsize$\pm 15.7$} & $64.0$ {\scriptsize$\pm 15.7$} & $63.4$ {\scriptsize$\pm 15.2$} \\
\midrule
\multirow{3}{*}{TabPFN-3.5} & Random & $60.4$ {\scriptsize$\pm 15.5$} & $60.1$ {\scriptsize$\pm 15.2$} & $59.8$ {\scriptsize$\pm 15.1$} & $57.9$ {\scriptsize$\pm 14.3$} \\
& TabICLv2 & $60.4$ {\scriptsize$\pm 15.5$} & $\mathbf{64.1}$ {\scriptsize$\pm 15.9$} & $\mathbf{64.5}$ {\scriptsize$\pm 15.9$} & $\mathbf{64.6}$ {\scriptsize$\pm 15.5$} \\
& TabDPT & $60.4$ {\scriptsize$\pm 15.5$} & $63.0$ {\scriptsize$\pm 16.0$} & $63.8$ {\scriptsize$\pm 16.0$} & $63.1$ {\scriptsize$\pm 15.5$} \\
\bottomrule
\end{tabular}
\end{table}

Table~\ref{tab:transfer} demonstrates that attributions computed on one backbone transfer effectively to curate demonstrations for unseen targets (\texttt{TabPFN-3} and \texttt{TabPFN-3.5}). At 50\% removal using transferred scores from either \texttt{TabICLv2} or \texttt{TabDPT}, balanced accuracy reaches approximately  64\%, exceeding full-context baselines by over four points and substantially outperforming random removal. Transfer is robust across source models, with \texttt{TabDPT} performing comparably to the stronger \texttt{TabICLv2}.



\subsection{TDA for TFM Active Learning}
\paragraph{\method{}-based Acquisition Strategy.} We finally investigate whether \method\ can guide data acquisition in pool-based active learning with TabICLv2, on the same training/validation/test splits as the curation experiment. For each dataset, the pool $\mathcal P$
is the training split, drawn anew in each repetition, for larger training sets.
The labeled context $\mathcal D_0$ starts with $n_0 = 32$ demonstrations of $\mathcal P$, one per class and the others drawn uniformly,
and the rest of $\mathcal P$ forms the unlabeled pool $\mathcal U_0$. At each round $t$, a strategy selects a batch $\mathcal B_t$ of
$b_t$ candidates from $\mathcal U_t$, whose labels are revealed and which join the context: $\mathcal D_{t+1} = \mathcal D_t \cup
\mathcal B_t$. The batch grows with the context, $b_t = \max\bigl(4, 2^{\lfloor \log_2 (|\mathcal D_t|/8) \rfloor}\bigr)$, and
acquisition stops once half of the pool is labeled. \method\ acquires the candidates whose addition would most reduce the surrogate loss on a labeled validation set $\mathcal V$. At each
round, the representations of the context, the candidates and $\mathcal V$ are read in a single forward pass with $\mathcal D_t$ as
context and held fixed (Equation~\ref{ticda:fixed}), and the surrogate of Equation~\ref{ticda:fit} is fitted on the context with $\lambda = 10$. Adding a candidate $x$ with label $c$ to the
context changes this surrogate in closed form (Equation~\ref{ticda:finite-parameters} with $t = -1$): $\widehat B(+x, c) = \widehat B - \frac{1}{1 + h_x}\, H^{-1} m_x \bigl(\widehat B^\top m_x - e_c\bigr)^\top$ with $h_x = m_x^\top H^{-1} m_x$ where $e_c$ is the one-hot vector of class $c$. \method\ scores $x$ under its most favorable label: $\mathrm{score}(x) = \max_{c} \Bigl[ L_{\mathcal V}\bigl(\widehat B\bigr) - L_{\mathcal V}\bigl(\widehat B(+x, c)\bigr) \Bigr]$, with $L_{\mathcal V}(B) = \frac{1}{K} \sum_{k=1}^{K} \frac{1}{|\mathcal V_k|} \sum_{q \in \mathcal V_k} \ell\bigl(B^\top m_q, y_q\bigr)$, where $\mathcal V_k$ holds the validation examples of class $k$, so that every class weighs equally, as in balanced accuracy. Each
batch is built greedily: after each selection, $H$ is updated with the representation of the selected candidate, a rank-one update
that requires no label, and the remaining candidates are rescored; $\widehat B$ is refitted once the labels of the batch are revealed.
Because early contexts contain far fewer demonstrations than the $d = 512$ dimensions of the representations, the representations are
first projected onto $d_t = \min\bigl(512, \max(4, \lfloor |\mathcal D_t|/5 \rfloor)\bigr)$ Gaussian random directions, redrawn at
every round, so that the surrogate is fitted with about five demonstrations per dimension; as TabICL averages eight ensemble views, one
surrogate is fitted per view and their predictions are averaged. $\mathcal V$ is a class-stratified sample of the validation split,
drawn once per repetition, that holds $20\%$ of the final budget (at least 32 examples); it is never added to the context, is not
counted in the labeling budget, and is not used by the baselines.

\paragraph{Experimental Protocol.}

We compare \method\ with random sampling, which draws the $b_t$ candidates uniformly, and with uncertainty sampling~\citep{lewis1994sequential}, a strong baseline that selects the
$b_t$ candidates on which the TFM is least certain. We name this baseline \textit{entropy}, since
it measures this uncertainty by the predictive entropy $-\sum_{k=1}^{K} p_k(x) \log p_k(x)$, where $p(x)$ is the class distribution
predicted by the TFM given $\mathcal D_t$. All strategies share the initial contexts, pools, test examples and batch sizes. After each
round, we measure the balanced accuracy on the test split, and summarize each learning curve by its normalized area under the
curve (AULC): the average balanced accuracy over $\log_2 |\mathcal D_t|$, from 32 labels to the end of the budget or to 512 labels.
Results are averaged over $R = 5$ repetitions, which differ in the initial context, the pool subsample and $\mathcal V$, on 27
datasets: 24 from TabArena, jannis~\citep{grinsztajn2022tree},
higgs~\citep{baldi2014searching} and otto~\citep{purucker2026beyond}. 

\paragraph{Experimental Results.}

\begin{figure*}[t!]
\noindent
\begin{minipage}[b]{0.53\textwidth}
\centering
\captionsetup{position=top}
\captionof{table}{Active learning: normalized area under the balanced-accuracy curve ($\times 100$, mean $\pm$ std over datasets). Bold: best; underlined: second; $^{*}$: paired $t$-test against the runner-up, $p<.01$.}
\label{tab:al}
\small
\setlength{\tabcolsep}{3.5pt}
\begin{tabular}[b]{lcc}
\toprule
 & Full budget & $\leq 512$ labels \\
 & (27 datasets) & (23 datasets) \\
\midrule
Random & 66.23 {\scriptsize $\pm$14.17} & 66.38 {\scriptsize $\pm$11.72} \\
\texttt{entropy} & \underline{67.56} {\scriptsize $\pm$14.95} & \underline{67.51} {\scriptsize $\pm$12.66} \\
\method & \textbf{68.35}$^{*}$ {\scriptsize $\pm$14.74} & \textbf{69.02}$^{*}$ {\scriptsize $\pm$12.09} \\
\bottomrule
\end{tabular}
\end{minipage}\hfill
\begin{minipage}[b]{0.43\textwidth}
\centering
\includegraphics[width=\linewidth]{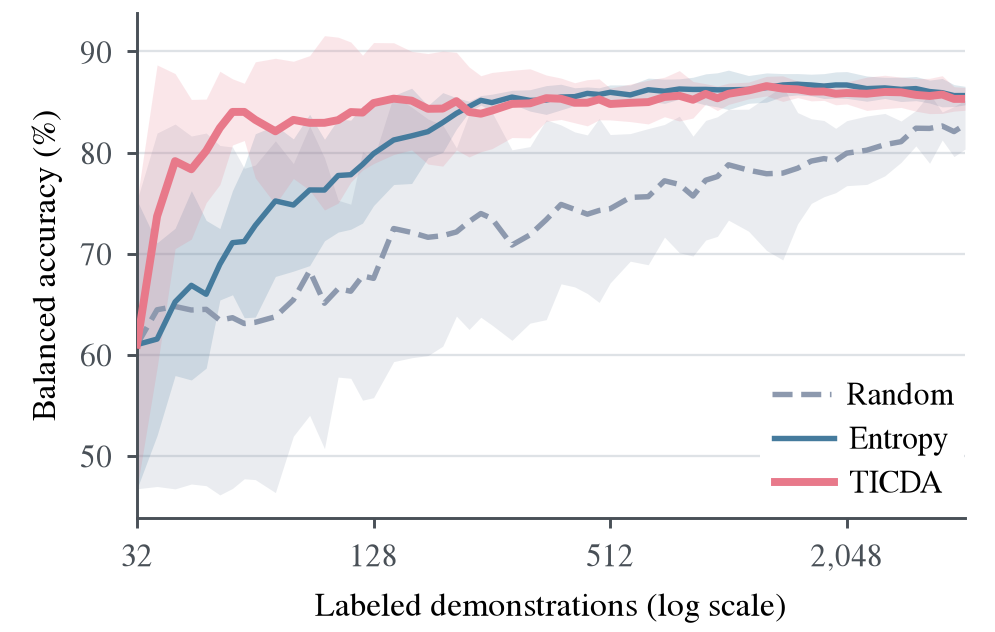}
\captionof{figure}{Balanced accuracy evolution on APSFailure against the number of labeled demonstrations.}
\label{fig:al}
\end{minipage}
\medskip
\end{figure*}

Table~\ref{tab:al} reports the aggregated results and Figure~\ref{fig:al} shows
an example of a learning curve on the APSFailure Dataset from TabArena. \method\ achieves the best average AULC across both the full budget (68.35\%) and the data-constrained regime with $\leq 512$ labels (69.02\%), outperforming entropy sampling and random baselines.




\section{Related Work}
\label{sec:related-work}

Training data attribution (TDA) methods attribute model performance to individual training points, grouped into three categories~\citep{hammoudeh2024training}: (1) \emph{resampling-based} methods, which retrain models on perturbed subsets; (2) \emph{gradient-based} methods, which approximate influence from gradients; and (3) \emph{training-dynamics} methods, which track influence across checkpoints. Their forms depend on the training regime: classical training (where point effects are absorbed into weights) versus in-context learning (where examples influence inference directly). We present in the following on resampling-based and gradient-based approaches, the only families developed for both regimes~\citep{deng2025survey}.


\paragraph{Resampling-based Approaches.}


Resampling-based approaches measure how a prediction or the model's performance changes under finite perturbations of the training set, removing or reweighting individual points, mirroring perturbation-based feature attribution such as \texttt{SHAP}~\citep{lundberg2017shap}. The elementary case is leave-one-out (LOO) retraining, formalized for linear regression by Cook's distance~\citep{cook1977detection}: a single point is removed and the model refit. \texttt{Data Shapley}~\citep{ghorbani2019datashapley} values each example by its marginal contribution to performance, at the cost of repeatedly retraining the model. \texttt{In-Run Data Shapley}~\citep{wang2025data} makes this affordable by defining a variant of these values for one specific training run, which it computes from the gradients of that run. \texttt{Data Shapley} extends naturally to the ICL setting, where the demonstrations play the role of the training set. \texttt{DemoShapley}~\citep{xie2025demoshapley} values each demonstration in the prompt used for Large Language Models (LLMs) by its marginal contribution across resampled prompt permutations. While this approach enables to detect high value data to improve LLM performance, this is only affordable for the few-shot LLM setting it targets, but not for the long contexts of TFMs, where the subsets grow combinatorially and each needs a full forward pass. 

\paragraph{Gradient-based Approaches.}


Gradient-based approaches approximate a training point's effect through the model's gradients, without retraining. Influence functions \citep{koh2017influence} provide a first-order Taylor approximation to
\texttt{LOO} retraining, approximating the effect of infinitesimally upweighting a
training point on the loss at a test point, which is the formulation of
Equation~\ref{eq:influence} from which \method\ is derived. The Hessian being computationally expensive to compute, several follow-up works have been proposed to make it tractable
for deep models by either estimating the Hessian more efficiently
\citep{schioppa2022scaling,kwon2024datainf,wang2025better} or by tracing gradients along the training run~\citep{pruthi2020estimating}. However, these approaches do not transfer naturally to ICL demonstrations, since a demonstration updates no model parameter. \texttt{DETAIL}~\citep{zhou2024detail} addresses the aforementioned limitations for Large Language Models (LLMs) and is the method conceptually closest to \method. Building on the view that transformers implement an internal optimizer during ICL~\citep{von2023transformers}, it models this learner as a linear ridge probe on the hidden states of the demonstrations and applies Equation~\ref{eq:influence} to this probe rather than to the model's weights, a surrogate view we share. Its motivation, design, and interpretation, however, do not transfer to TFMs, because \texttt{DETAIL} is built for LLM prompts. The two methods differ in three respects: the regime in which the probe is fitted, since LLM prompts hold few demonstrations in a high-dimensional space whereas TFM contexts hold many rows in a low-dimensional one; the perturbation each score differentiates; and the treatment of contextual representations under a single forward pass. We defer the detailed analysis to Appendix~\ref{app:ticda_vs_detail}.

\section{Discussion and Conclusion}
\label{sec:discussion}

\method\ rests on the fixed-representation approximation of Equation~\ref{ticda:fixed}: removing a demonstration is assumed to leave the representations of the remaining rows and the query unchanged, so that only the surrogate fit is refitted. 
Appendix~\ref{ticda:app-fixed-measure} shows that this is well justified when a single demonstration is a small fraction of a large context. 
Our faithfulness evaluation compares \method\ with exhaustive \texttt{LOO} on all 38 TabArena datasets, but only with contexts of at most \(4{,}096\) demonstrations. Longer contexts, which \method\ is meant to serve, remain untested against this reference, since exhaustive deletion requires one forward pass of the TFM per demonstration.
We see \method\ as a promising methodology to optimize context under ICL settings that could be integrated into other existing framework~\citep{bhan2024self, yang2023representative, xu2023knn} carefully targeting ICL examples. Finally, extending \method\ beyond the fixed-representation approximation and to regression and multimodal settings are natural directions for future work.

In this paper we introduced \method, a gradient-based data attribution method for tabular foundation models that estimates the influence of every in-context demonstration in a single forward pass. \method\ overcomes the central obstacle to applying influence functions under in-context
learning, the absence of a parameter update, by fitting a ridge surrogate on the frozen row representations and holding them fixed under
demonstration removal, yielding closed-form influence scores at negligible cost. Across 38 TabArena datasets and two backbones, \method\ is the only method that pairs top-tier faithfulness and labeling-error detection with
a runtime well below \texttt{LOO}, \texttt{DemoShapley}, and \texttt{IG}. Its curation preserves or improves
model performance while discarding up to half of the demonstrations, and the scores transfer across TFMs, allowing attribution to be computed once and
amortized. 



\section*{AI use statement}

Generative AI tools have been used to polish the paper and help in generating tables and figures. These tools also helped in reviewing the manuscript to detect typos and inconsistencies. We also used AI tools to accelerate code production. We take full responsibility for the final content.

\bibliography{ref}
\bibliographystyle{iclr2027_conference}
\clearpage
\appendix
\section{Scientific Libraries}
\label{ticda:sci_lib}

We used several open-source libraries in this work: \verb|pytorch|~\citep{paszke2017automatic}, HuggingFace
\verb|transformers|~\citep{wolf2019huggingface} and \verb|sklearn|~\citep{pedregosa2011scikit}. We will make our code public
upon acceptance.

\section{Derivation of the influence scores}
\label{ticda:derivation}

This appendix derives the scores of Section~\ref{ticda:method} and
relates them to the exact quantities that they approximate. We first
prove the closed form of the influence score
(Equation~\ref{ticda:support}), then give the exact effect of deleting a
demonstration when the representations are held fixed, which shows what
a first-order score leaves out, state what the fixed-representation simplification of Equation~\ref{ticda:fixed} itself neglects, and finally measure how much the representations change when a demonstration is removed. 

Let $a_j=m_jr_j^{\top}\in\mathbb{R}^{d\times K}$
denote the gradient of the loss $\ell_j$ of row $j$ at $\widehat{B}$,
and $\langle U,V\rangle_F=\operatorname{tr}(U^{\top}V)$ denote the Frobenius
inner product of two matrices.

\subsection{First-order influence}
\label{ticda:app-first-order}

To obtain Equation~\ref{ticda:support}, we need the derivative of the
fitted coefficients with respect to the weight of a demonstration, since
the query loss depends on this weight only through the coefficients. We
follow the derivation of influence functions \citep{koh2017influence},
every step of which is available in closed form for a ridge surrogate.

The perturbed objective of Equation~\ref{ticda:perturbation} is a ridge
regression in which demonstration $i$ has weight $1+\epsilon$, so that
its minimizer $\widehat{B}_\epsilon$ satisfies the normal equations:
\begin{equation*}
 \big(H+\epsilon\,m_im_i^{\top}\big)\widehat{B}_\epsilon
 =M^{\top}Y+\epsilon\,m_iy_i^{\top},
\end{equation*}
where $H=M^{\top}M+\lambda I_d$ is positive definite for $\lambda>0$.
Differentiating these equations at $\epsilon=0$ and using
$r_i=\widehat{B}^{\top}m_i-y_i$ gives
\begin{equation}
 \left.\frac{d\widehat{B}_\epsilon}{d\epsilon}\right|_{\epsilon=0}
 =-H^{-1}m_ir_i^{\top}=-H^{-1}a_i.
 \label{ticda:parameter-derivative}
\end{equation}
The query loss depends on $\epsilon$ only through the coefficients, and
its gradient with respect to $B$ at $\widehat{B}$ is $a_q$, so that the
chain rule gives
\begin{equation*}
 \left.\frac{d\,\ell_q(\widehat{B}_\epsilon)}{d\epsilon}\right|_{\epsilon=0}
 =\Big\langle a_q,\,
   \left.\frac{d\widehat{B}_\epsilon}{d\epsilon}\right|_{\epsilon=0}
  \Big\rangle_F
 =-\big\langle a_q,\,H^{-1}a_i\big\rangle_F .
\end{equation*}
Equation~\ref{ticda:support} defines $s_{iq}$ as the opposite of this
derivative, and the identity
$\langle m_qr_q^{\top},H^{-1}m_ir_i^{\top}\rangle_F
=(m_q^{\top}H^{-1}m_i)(r_q^{\top}r_i)$ gives its closed form. The matrix $H$ is the Hessian of
the surrogate objective with respect to each of the $K$ columns of $B$,
so that the inverse Hessian of Equation~\ref{eq:influence} reduces to
$K$ copies of the inverse of a $d\times d$ matrix. No Hessian over the
parameters of the TFM is therefore involved. Equation~\ref{eq:influence} is
written for an objective that averages the losses, whereas
Equation~\ref{ticda:objective} sums them. The two conventions define the
same surrogate when the averaged objective uses the coefficient
$\lambda/n$, and they give scores that differ by the constant factor
$n$, which leaves every ranking unchanged.

Taking the demonstration itself as the query gives the self-influence equation introduced in Section~\ref{ticda:influence}, $T_i=h_i\lVert r_i\rVert_2^{2}$, where
\begin{equation}
 h_i=m_i^{\top}H^{-1}m_i
 \label{ticda:leverage}
\end{equation}
is the ridge leverage of demonstration $i$, that is, the sensitivity of
the prediction of the surrogate for this demonstration to its own label.
Since $H-m_im_i^{\top}$ remains positive definite, the leverages satisfy
$0\le h_i<1$, and their sum is
$\sum_{i}h_i=d-\lambda\operatorname{tr}(H^{-1})<d$, so that the mean
leverage of a context is below $d/n$. In a tabular context, where $n$ exceeds $d$ by an order of magnitude, the leverage therefore multiplies the squared residual by a small factor on average, and this factor is
largest for the demonstrations whose representation lies in a direction that few other demonstrations occupy. The self influence equation introduced in Section~\ref{ticda:influence} uses, on both sides of the product, the representation that the TFM computes for row $i$ in its role of demonstration. Encoding the same row as a query may give a different vector, since a TFM receives a demonstration with its label and a query without it; the corresponding score would require an additional forward pass, and we do not use it.

\subsection{Exact deletion at fixed representations}
\label{ticda:app-exact}

The score of Equation~\ref{ticda:support} is a derivative, whereas
deleting a demonstration changes its weight from one to zero. That
raises the question of how far a first-order score is from the effect of
an actual deletion. When the representations are held fixed and the surrogate is a ridge
regression on a fixed design, this question has a closed-form answer. We derive this exact effect below, first on a query and then on the demonstration itself, and show that it differs from the first-order score by corrections that vanish with the leverage, so that the two nearly coincide in a tabular context. This exact effect also serves as a reference for \TICDA{} in Section~\ref{sec:experimental_setting}.

Let $\widehat{B}(t)$ minimize $J(B)-t\,\ell_i(B)$ for
$0\le t\le1$, so that $t=1$ corresponds to the context without
demonstration $i$. Since this objective differs from $J$ by a term of
rank one, the Sherman--Morrison identity \citep{sherman1950adjustment}
gives
\begin{equation}
 \widehat{B}(t)-\widehat{B}
 =\frac{t}{1-t\,h_i}\,H^{-1}m_ir_i^{\top},
 \label{ticda:finite-parameters}
\end{equation}
whose derivative at $t=0$ is the opposite of
Equation~\ref{ticda:parameter-derivative}. Writing
$a_{qi}=m_q^{\top}H^{-1}m_i$, the deletion therefore changes the loss of
the surrogate on query $q$ by
\begin{equation}
 \Delta^{\mathrm{F}}_{iq}
 =\frac{a_{qi}}{1-h_i}\,r_q^{\top}r_i
 +\frac{a_{qi}^{2}}{2(1-h_i)^{2}}\,\lVert r_i\rVert_2^{2},
 \label{ticda:finite-query}
\end{equation}
where the first term is the score $s_{iq}$ divided by $1-h_i$, and the
second term is of second order in the leverages, since
$\lvert a_{qi}\rvert\le\sqrt{h_ih_q}$ with
$h_q=m_q^{\top}H^{-1}m_q$. For the demonstration itself,
$a_{ii}=h_i$, the residual after deletion is $r_i/(1-h_i)$, and the
change of its loss is
\begin{equation}
 \Delta^{\mathrm{F}}_{ii}
 =\frac12\lVert r_i\rVert_2^{2}\Big[(1-h_i)^{-2}-1\Big]
 =T_i\,\frac{1-h_i/2}{(1-h_i)^{2}},
 \label{ticda:finite-self}
\end{equation}
so that the exact effect and the self-influence differ by a factor that
depends on the leverage alone and that tends to one as the leverage
tends to zero. Since the mean leverage is below $d/n$, this factor is
close to one for most demonstrations of a tabular context, which is the
regime in which the first-order score is accurate. It is not accurate in
a context that has fewer demonstrations than dimensions, where the
leverages approach one. Equation~\ref{ticda:finite-query} is the exact
fixed-representation deletion of Section~\ref{sec:experimental_setting}. It
requires no refit, since it reuses the inverse of $H$, the residuals and
the leverages of the single fit.

\subsection{What the fixed-representation simplification neglects}
\label{ticda:recomputation}

The two previous results hold for representations that do not respond to
the perturbation, which is the simplification of
Equation~\ref{ticda:fixed}. To state what this simplification leaves
out, suppose that the context varies along a differentiable path indexed
by $\epsilon$, so that the coefficients and the representations both
respond. The chain rule then gives
\begin{equation}
 \frac{d\ell_q}{d\epsilon}
 =r_q^{\top}\dot{B}^{\top}m_q+r_q^{\top}\widehat{B}^{\top}\dot{m}_q,
 \label{ticda:chain-rule}
\end{equation}
where $\dot{B}$ and $\dot{m}_q$ denote derivatives with respect to
$\epsilon$. The second term is the response of the representation of the
query, and the derivative $\dot{B}$ of the first term contains, in
addition to Equation~\ref{ticda:parameter-derivative}, the response of
every row of $M$. Equation~\ref{ticda:fixed} sets these two responses to
zero and keeps Equation~\ref{ticda:parameter-derivative} alone.

The finite counterpart of this statement is the recomputed deletion
$\Delta^{\mathrm{R}}_{iq}$ of Equation~\ref{ticda:recomputed}, in which
we remove demonstration $i$ from the context, let the TFM encode the
remaining demonstrations and the query again, and refit the surrogate
with the same regularization. The difference between
$\Delta^{\mathrm{R}}_{iq}$ and $\Delta^{\mathrm{F}}_{iq}$ measures the
effect of Equation~\ref{ticda:fixed} alone, provided that the two
quantities share the queries, their role and their labels. The
difference between $\Delta^{\mathrm{R}}_{iq}$ and the leave-one-out
effect $\Delta^{\mathrm{TFM}}_{iq}$ of Equation~\ref{ticda:loo} measures
in turn what the surrogate does not capture, since both quantities
encode the context again and differ only in the loss that they read.
Both require one forward pass of the TFM per deleted demonstration,
which is the cost that \TICDA{} avoids.

\section{Relation to \texttt{DETAIL}: the role of the ridge penalty}
\label{app:ticda_vs_detail}

This appendix explains how the score of  \TICDA{}
differs from that of \DETAIL{} \citep{zhou2024detail}. Although the two methods are designed for different domains of applications, hence their motivations and the justifications of the approximation/simplification, as explained in Section \ref{sec:related-work}, they both rely on the same
surrogate. Both methods fit the ridge regression of
Equation~\ref{ticda:objective}, so they share the coefficients
$\widehat{B}$ and the matrix $H$, and both apply
Equation~\ref{eq:influence} to it. The main difference in the design is in the construction of each demonstration's loss: for \TICDA{}, it is the squared error $\ell_i(B)$ of the demonstration, whereas \DETAIL{}
adds the ridge penalty to it, in full and not as a share divided among
the $n$ demonstrations:
\begin{equation*}
 p_i(B)=\ell_i(B)+\frac{\lambda}{2}\lVert B\rVert_F^{2},
 \label{ticda:detail-loss}
\end{equation*}
where $\lVert\cdot\rVert_F$ is the Frobenius norm, so that the second
term is the penalty of $J$ itself. We show below what this choice changes, first for the question that each
score answers, then for the gradient of a demonstration, and finally for
the score itself.

An influence function estimates, to first order, what happens to a model
when the loss of one example is removed from its training objective. 
What is removed therefore depends on what a method calls the loss of a
demonstration.  To compare the two cases, we write the objective of
Equation~\ref{ticda:objective} as a sum over the demonstrations
\begin{equation*}
    J(B)=\sum_j\ell_j(B)+\tfrac{\lambda}{2}\lVert B\rVert_F^{2}
\end{equation*}
For \TICDA{}, the loss of demonstration $i$ is $\ell_i$, and removing it
from $J$ gives
\begin{equation*}
 J(B)-\ell_i(B)
 =\sum_{j\ne i}\ell_j(B)+\frac{\lambda}{2}\lVert B\rVert_F^{2},
\end{equation*}
i.e., the other demonstrations with the same penalty. 
\TICDA{} therefore estimates the effect of fitting the surrogate without
demonstration $i$ and with the same regularization, which is
leave-one-out (Equation~\ref{ticda:loo}) applied to the surrogate
instead of the TFM.

For \DETAIL{}, the loss of
demonstration $i$ is $p_i$, and removing it from $J$ gives
\begin{equation*}
 J(B)-p_i(B)=\sum_{j\ne i}\ell_j(B),
\end{equation*}
i.e., the other demonstrations with no penalty at all, since $J$
contains the penalty only once. \DETAIL{} therefore estimates the joint
effect of removing the demonstration and of removing the regularizer.

The same difference appears in the gradients from which the two scores
are computed. With the notation of Appendix~\ref{ticda:derivation}, the
gradient of $\ell_i$ at $\widehat{B}$ is $a_i=m_ir_i^{\top}$, for \DETAIL{} the
gradient at demonstration $i$ is
\begin{equation*}
 g_i=a_i+\lambda\widehat{B},
\end{equation*}
\DETAIL{} adds the same matrix $\lambda\widehat{B}$ to the
gradient of every demonstration, and to that of the query.

To see what
this added matrix represents, we use the fact that $\widehat{B}$
minimizes $J$. Since $J$ is the sum of the losses $\ell_j$ and of
the penalty, its gradient at $\widehat{B}$ is
$\sum_j a_j+\lambda\widehat{B}$, and this gradient is zero because
$\widehat{B}$ minimizes $J$, the matrix that \DETAIL{} adds is therefore
minus the sum of the gradients of all the demonstrations:
\begin{equation*}
 \lambda\widehat{B}=-\sum_{j}a_j .
\end{equation*}
Substituting this expression into $g_i=a_i+\lambda\widehat{B}$ cancels the
term $a_i$ and gives
\begin{equation}
 g_i=-\sum_{j\ne i}a_j .
 \label{ticda:detail-identity}
\end{equation}

The gradient that \DETAIL{} attributes to demonstration $i$ is thus the
sum of the gradients of all the other demonstrations, with the opposite
sign. For example, in a context of $1{,}000$ demonstrations, the
gradients that it attributes to two demonstrations $i$ and $j$ are two sums of $999$
terms $k$, $k \neq i$ or $k \neq j$, respectively. $998$ components of these two sums are the same. Hence, what \DETAIL{} adds is a term common to all
demonstrations.

Finally, consider the score that each method gives to a demonstration
for its own loss, which we call its self-score. Up to a constant factor, the self-score of demonstration $i$ is the quadratic form $\langle g_i,H^{-1}g_i\rangle_F$ of the gradient $g_i$
that the method attributes to $i$, where
$\langle\cdot,\cdot\rangle_F$ is the Frobenius inner product of
Appendix~\ref{ticda:derivation}. For \TICDA{}, the gradient is $a_i$ and
this quadratic form is the self-influence
$T_i=\langle a_i,H^{-1}a_i\rangle_F$
(Appendix~\ref{ticda:app-first-order}). Replacing $a_i$ by
$a_i+\lambda\widehat{B}$, like in \DETAIL{}, expands this quadratic form as the square of a
sum, which produces two additional terms:
\begin{equation}
 \big\langle g_i,\,H^{-1}g_i\big\rangle_F
 =T_i
 +\underbrace{2\lambda\,\big\langle a_i,\,H^{-1}\widehat{B}\big\rangle_F}_{\text{cross term }X_i}
 +\underbrace{\lambda^{2}\big\langle\widehat{B},\,H^{-1}\widehat{B}\big\rangle_F}_{\text{constant }C}.
 \label{ticda:detail-decomposition}
\end{equation}
We denote the middle term by $X_i$ and last term by $C$. $C$ is the same for all demonstrations,
so that it has no effect on their ranking. The cross term $X_i$ depends on the demonstration,
and it has a direct reading. As in Equation~\ref{ticda:perturbation}, let $\epsilon$ be the
additional weight given to demonstration $i$ and $\widehat{B}_\epsilon$
the corresponding minimizer. By Equation~\ref{ticda:parameter-derivative} and the symmetry
of $H^{-1}$, the penalty term of this fit responds to the weight as
\begin{equation*}
 \left.\frac{d}{d\epsilon}\,
   \frac{\lambda}{2}\lVert\widehat{B}_\epsilon\rVert_F^{2}
 \right|_{\epsilon=0}
 =\lambda\Big\langle\widehat{B},\,
   \left.\frac{d\widehat{B}_\epsilon}{d\epsilon}\right|_{\epsilon=0}
  \Big\rangle_F
 =-\lambda\,\big\langle a_i,\,H^{-1}\widehat{B}\big\rangle_F ,
\end{equation*}
where the first equality is the derivative of a squared norm and the second substitutes Equation~\ref{ticda:parameter-derivative}. The cross term $2\lambda\langle a_i,H^{-1}\widehat{B}\rangle_F$ is thus equal to minus twice this derivative. It is positive when the derivative is negative, which happens for a demonstration whose weight reduces the norm of the coefficients, and it is negative otherwise. Since it can take either sign, it can change the ranking given by $T_i$. The two scores thus answer different questions.

\subsection{Empirical check of the fixed-representation simplification}
\label{ticda:app-fixed-measure}

The simplification of Equation~\ref{ticda:fixed} assumes that removing a demonstration leaves the representations of the remaining demonstrations and of the query unchanged, whereas a TFM computes each of these representations from the whole context. To see how far this assumption is from what the TFM actually does, and how this gap depends on the size of the context, we remove a demonstration, let the TFM encode the reduced context again, and compare every representation with the one computed from the full context. For a removed demonstration $i$, we measure the relative change in the representation of row $j$, either a remaining demonstration or a query:
\begin{equation}
  \rho_{ij} = \frac{\lVert \phi(x_j \mid \mathcal{D}_{\setminus i}) - \phi(x_j \mid \mathcal{D}) \rVert_2}{\lVert \phi(x_j \mid \mathcal{D}) \rVert_2},
  \label{ticda:rel-change}
\end{equation}
where $\lVert\cdot\rVert_2$ is the Euclidean norm. 

We measure this ratio for TabICL, with the same pre-ICL representations as in Section~\ref{sec:experimental_protocol}. We use four TabArena datasets that are large enough for contexts of 8,192 demonstrations: GiveMeSomeCredit, SDSS17, bank-marketing and HR\_Analytics. For each dataset, we draw 50 queries from the test split and random contexts of 512, 1,024, 2,048, 4,096 and 8,192 demonstrations from the training split. As in Section~\ref{sec:experimental_protocol}, we corrupt $20\%$ of the labels in each context. We then pick 16 demonstrations, half of them mislabeled, and delete each of them separately: we remove that single demonstration, encode the remaining context, and put it back before the next deletion. The 16 demonstrations that we delete are in every context. We repeat this for two random draws at every size per dataset.

\begin{table}[h]
  \centering
  \caption{Relative change $\rho_{ij}$ of the TabICL representations when one demonstration is removed (Equation~\ref{ticda:rel-change}), in percent of the norm of the representation, for contexts of increasing size. For each dataset, we take the median of $\rho_{ij}$ over all pairs of a removed demonstration $i$ and a row $j$, and we report the mean $\pm$ standard deviation of this median over the four datasets. A value of zero would mean that the representations do not change.}
  \label{tab:rel-change}
  \begin{tabular}{rcc}
    \toprule
    Demonstrations in the context & Remaining demonstrations & Queries \\
    \midrule
    512   & 2.90 {\scriptsize $\pm$0.60} & 2.93 {\scriptsize $\pm$0.62} \\
    1,024 & 1.73 {\scriptsize $\pm$0.61} & 1.71 {\scriptsize $\pm$0.61} \\
    2,048 & 0.89 {\scriptsize $\pm$0.24} & 0.88 {\scriptsize $\pm$0.25} \\
    4,096 & 0.48 {\scriptsize $\pm$0.14} & 0.48 {\scriptsize $\pm$0.15} \\
    8,192 & 0.24 {\scriptsize $\pm$0.07} & 0.24 {\scriptsize $\pm$0.08} \\
    \bottomrule
  \end{tabular}
\end{table}

\begin{figure}[h]
  \centering
  \includegraphics[width=0.7\linewidth]{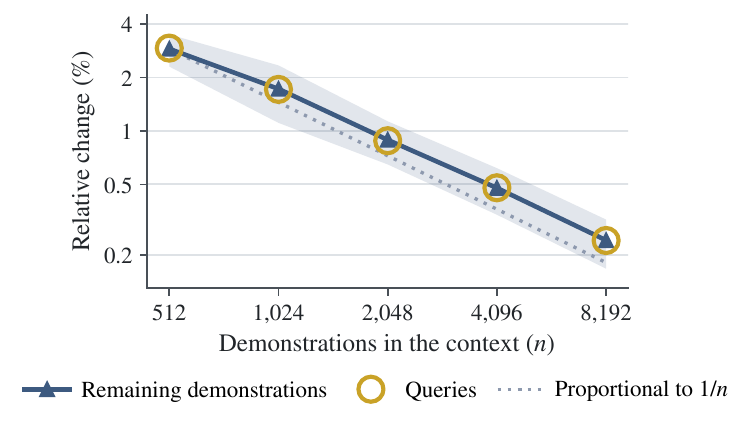}
  \caption{Relative change $\rho_{ij}$ of the TabICL representations when one demonstration is removed, against the number $n$ of demonstrations in the context, on logarithmic axes. The markers show the mean over the four datasets of the per-dataset median, as in Table~\ref{tab:rel-change}, and the band one standard deviation for the remaining demonstrations. The change of the queries coincides with that of the remaining demonstrations, and both follow the dotted line, which decreases as $1/n$.}
  \label{fig:rel-change}
\end{figure}

Table~\ref{tab:rel-change} and Figure~\ref{fig:rel-change} show that the representations do change after a deletion, but by a small amount that shrinks as the context grows: the median change roughly halves each time the number of demonstrations doubles, from about $3\%$ of the norm with 512 demonstrations to about $0.5\%$ with 4,096, the largest context of Section~\ref{sec:experimental_protocol}. This decrease is close to the rate $1/n$ at which the share of a single demonstration in the context falls, and it holds in each of the four datasets, for the remaining demonstrations as well as for the queries. The change is larger when the removed demonstration is itself mislabeled, by a factor of $1.6$ to $1.8$ at every context size. Equation~\ref{ticda:fixed} is therefore an approximation rather than an exact property of the TFM, and its error on each representation becomes small in the large contexts that TICDA is designed for. 

\section{Implementation Details}
\label{app:implementation_details}

In the following we give additional information about implementation of methods analyzed in the body of the paper. Integrated gradients (\texttt{IG})~\citep{sundararajan2017axiomatic} are computed with 32 steps to approximate gradients with Riemann sum on the straight line between instance to explain and baselines.

\texttt{DETAIL} is implemented based on hyperparameters provided by the paper~\citep{zhou2024detail}, with a regularization term at 10. 

\section{Additional Experiments}
\label{app:add_xp}

Below we show how the regularization term $\lambda$ impacts \method\ performance both in terms of labeling error detection and context curation. 

\begin{figure}[H]
    \centering
    \begin{subfigure}{\textwidth}
        \centering
        \includegraphics[width=\linewidth]{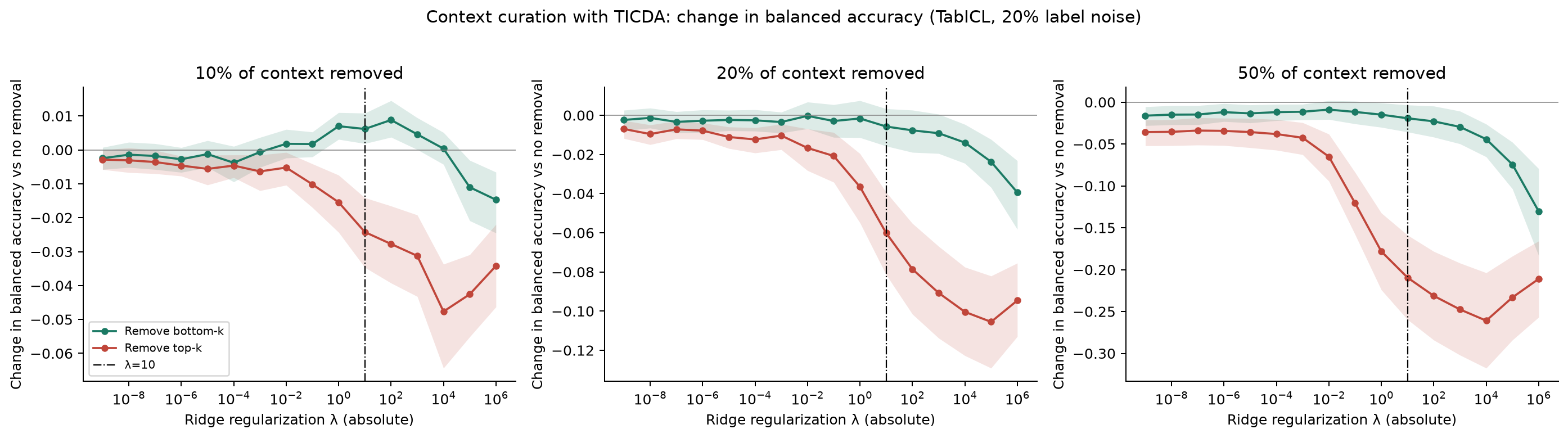}
        \caption{Context curation sensitivity to $\lambda$.}
        \label{fig:chart1}
    \end{subfigure}
    \\[1ex]
    \begin{subfigure}{\textwidth}
        \centering
        \includegraphics[width=0.5\linewidth]{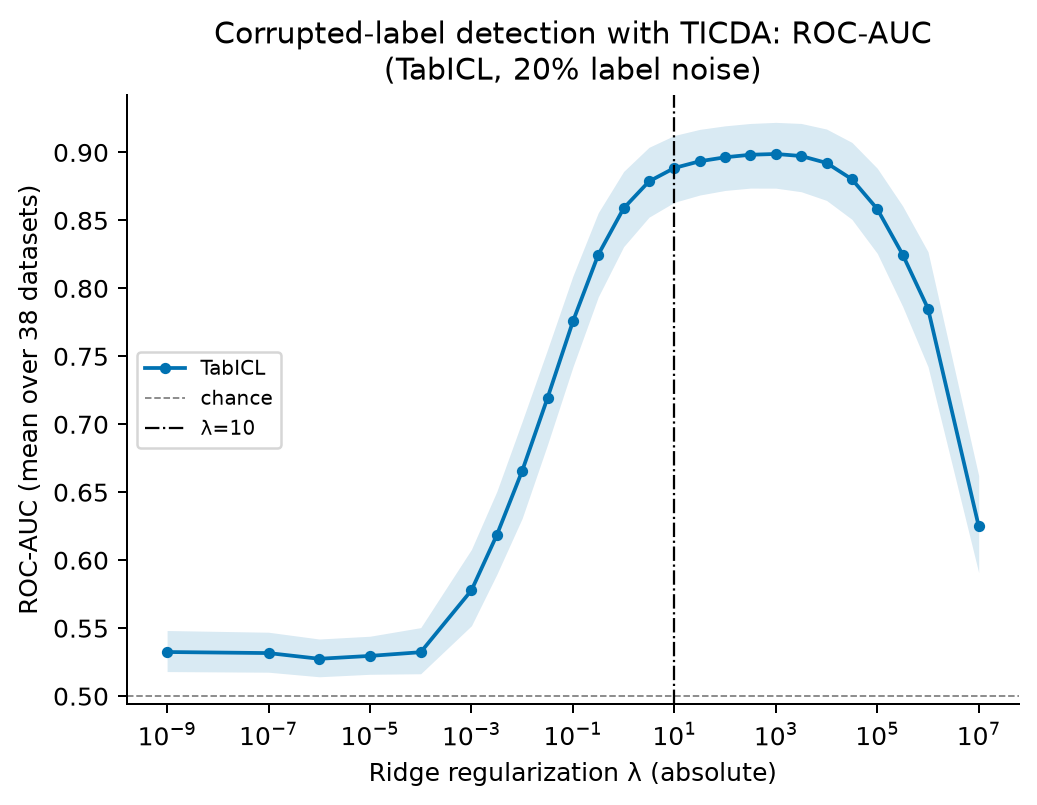}
        \caption{Labeling error detection sensitivity to $\lambda$.}
        \label{fig:chart2}
    \end{subfigure}
    \caption{\method\ sensitivity to $\lambda$ regularization.}
    \label{fig:both}
\end{figure}


\end{document}